%% file: main.tex
\documentclass[10pt,letterpaper,conference]{ieeeconf}
\usepackage{epsfig}
\usepackage{graphicx}
\usepackage{psfrag,graphicx,color}
\usepackage{amsmath}
\usepackage{algorithm2e}
\usepackage[mathscr]{eucal}
\usepackage{booktabs}

\IEEEoverridecommandlockouts
\usepackage[normalem]{ulem}
\usepackage{amsmath,amsfonts}
\usepackage{epsfig,psfrag,graphicx,color}
\input{bmit2e}

\input{mymacro}

\input{texdraw}

\definecolor{myorange}{rgb}{1,0.5,0}

\begin{document}

\title{\LARGE \bf Safety-related Tasks within the \\ Set-Based Task-Priority Inverse Kinematics Framework}
\author{Paolo Di Lillo, Filippo Arrichiello, Gianluca Antonelli, Stefano Chiaverini
\thanks{Authors are with the Department of Electrical and Information Engineering of the University of Cassino and Southern Lazio,
		Via G. Di Biasio 43, 03043 Cassino (FR), Italy
		}}

\maketitle

\begin{abstract}

In this paper we present a framework that allows the motion control of a robotic arm automatically handling different kinds of safety-related tasks. The developed controller is based on a Task-Priority Inverse Kinematics algorithm that allows the manipulator's motion while respecting constraints defined either in the joint or in the operational space in the form of equality-based or set-based tasks. This gives the possibility to define, among the others, tasks as joint-limits, obstacle avoidance or limiting the workspace in the operational space. Additionally, an algorithm for the real-time computation of the minimum distance between the manipulator and other objects in the environment using depth measurements has been implemented, effectively allowing obstacle avoidance tasks.  Experiments with a Jaco$^2$ manipulator, operating in an environment where an RGB-D sensor is used for the obstacles detection, show the effectiveness of the developed system.

\end{abstract}
\section{Introduction}
In the last years we are experiencing the spread of robots in several fields of the human life. Human-Robot collaboration, unthinkable until few years ago, now represents a growing topic of research in many areas such as industrial, medical and assistive and service robotics.
An explanatory example is the industrial field. During 70's and 80's robots were mainly used  for substituting humans in dirty, hard and repetitive jobs but they were forced to work in safety cages, preventing the cooperation between human
operators and robots for clear safety issues. During the years robots have gained more and more capabilities, due to several factors such as the 
falling of the sensor prices, the increase in computing power and the spread of the open-source development. All this resources allowed the spread of a new paradigm for the manufacturing industry based on the cooperation between human and robots \cite{lu2017industry}. In this perspective acquires a major importance the problem of the safety of the operations \cite{avanzini2014safety, zanchettin2016safety}.

Robotic manipulators are often required to perform tasks in the operational space, such as to move the end-effector at a certain position and/or orientation. However, a number of
additional tasks have to be taken into account while controlling the system in order to assure the safety and the effectiveness of the operation. The arm has to avoid obstacles, respect its mechanical joint limits, handle the occurrence of kinematic singularities. In the following, this kind of tasks will be called \textit{set-based}, because the control objective is to keep them in a certain set of values rather that a specific one. One of the first attempts to handle the obstacle avoidance task for a mobile robot is conducted in \cite{khatib1986} where the it is pushed away from the obstacle by defining a virtual force or potential field. This kind of approach presents drawbacks: it is not possible to set a 
minimum distance from the obstacle that the robot has to maintain, and an undesirable oscillating behavior of the system in presence of some kind of obstacles can occur \cite{koren1991potential}. More recently, a popular way to handle set-based tasks is to express the inverse kinematics problem in a sequence 
of QP (Quadratic Programming) problems \cite{kanoun_tro11, Mansard_tro2009}. This method requires the usage of iterative algorithms to solve the
optimization problems, and usually they are computationally heavy and slow.
In \cite{Simetti_jirs2016} set-based tasks are successfully added to a prioritized hierarchy and the transitions are handled by proper activation
functions that guarantee the smoothness of the output reference velocity. However, during the transitions, the strict priority order
among the tasks is lost, potentially leading to undesirable behaviors.

In this paper we present a system that allows the execution of the operational tasks, such as the control of the end-effector position and orientation, while all the safety-related ones such as the obstacle avoidance and the mechanical joint limits  are automatically handled by the system. 
Regarding the control algorithm, the key idea is to exploit the system redundancy. A robotic system is defined as redundant if it has more DOFs than those strictly needed for the accomplishment of a certain task. In this case, it is possible to perform multiple
tasks simultaneously \cite{chiaverini1997singularity}. The approach has been further extended to multiple prioritized tasks in \cite{AntArrChi_paladyn2010} and \cite{AntArrChi_cdc2008}, in which a priority order among the tasks can be fixed and the velocity contributions of the lower priority tasks that would conflict with higher priority ones are filtered. The outcomes of these works have been extended to handle also \textit{set-based} tasks \cite{MoeAnt2016frontiers},  \cite{arrichiello2017assistive}.
Regarding the obstacle avoidance task, we used a Kinect 2 sensor for monitoring the environment, exploiting the algorithm presented in \cite{flacco2015depth} for the detection of the closest obstacles using depth measurements. 

The paper is organized as follows:
Section \ref{sec:algorithm} describes the Multi-Task Inverse Kinematics Framework and the algorithm for handling the set-based tasks;
Section \ref{sec:distances} shows the algorithm for the obstacles detection using depth measurements;
in Section \ref{sec:experiments} experimental results are shown;
Section \ref{sec:conc} describes the conclusions and the future work.

\section{Multi-Task Priority Framework including set-based tasks}\label{sec:algorithm}
For a general robotic system with $n$ DOF (Degrees Of Freedom), the state is described by the joint values $\bfq = \left[q_1,q_2,\dots,q_n\right]^T\in \mathbb{R}^{\it{n}}$. Defining a \textit{task} as a generic $m$-dimensional control objective as a function of the system state $\boldsymbol{\sigma}(\bfq) \in \mathbb{R}^{\it{m}}$, the following differential relationship between the system velocity and the task-space velocity holds \cite{siciliano2010robotics}:

\begin{equation}
\dot\bfsigma(\bfq)=\bfJ(\bfq)\dot\bfq\;,
\end{equation}
where $\bfJ(q)=\frac{\partial \bfsigma(\bfq)}{\partial \bfq} \in \mathbb{R}^{\it{m\times n}}$ is the task Jacobian matrix, and $\dot\bfq$ is the joint velocity vector. The reference velocity that brings the task value $\bfsigma$ to a desired $\bfsigma_d$ can be computed resorting to the Closed-Loop Inverse-Kinematics algorithm \cite{chiaverini1997singularity}:
\begin{equation}\label{eq:diff}
\dot\bfq=\bfJ^\dagger(\dot\bfsigma_d+\bfK \tilde\bfsigma)\;,
\end{equation}
where $\bfK \in \mathbb{R}^{\it{m\times m}}$ is a positive-definite matrix of gains, $\tilde\bfsigma=\bfsigma_d-\bfsigma$ is the task error and
\begin{equation}
\bfJ^\dagger=\bfJ^T (\bfJ \bfJ^T)^{-1}
\end{equation}
is the Moore-Penrose psudoinverse of the Jacobian matrix $\bfJ$. 

If the system is redundant ($n>m$) it is possible to perform multiple tasks simultaneously, setting a priority level  among them and then projecting the velocity components coming from a lower priority task into the null space of the higher priority ones. In this way the fulfilment of the primary task is guaranteed. Thus for a prioritized hierarchy composed by $h$ tasks, the reference system velocity can be computed resorting to the \textit{Null-Space Based Inverse Kinematics control} \cite{AntArrChi_paladyn2010} \cite{AntArrChi_ISR08}:
\begin{equation}\label{eq:nsb}
\dot\bfq=\dot\bfq_1+\bfN_1 \dot\bfq_2+\dots+\bfN_{1,h-1} \dot\bfq_h\;,
\end{equation}
where $\dot\bfq_i$ is the reference velocity that fulfills the $i$-th task and $\bfN_{1,i}$ is the null space of the augmented Jacobian:
\begin{equation} \label{eq:jaco}
\bfJ_{1,i}=
\begin{bmatrix}
              \bfJ_1^T  &  \bfJ_2^T & \hdots & \bfJ_i^T
\end{bmatrix}^T
\end{equation}
computed as:
\begin{equation}
\bfN_{1,i} = \bfI-(\bfJ^\dagger \bfJ)^{-1}\;,
\end{equation}
where $\bfI$ is the $n$ by $n$ identity matrix.

The aforementioned framework has been developed to handle control objectives in which the goal is to bring the task value to a specific one, e.g. moving the arm end-effector to a target position. This kind of tasks are usually referred as \textit{equality-based}. However, several control objectives may require their value to lie in an interval, i.e. above a lower threshold and below an upper threshold. These are usually called \textit{set-based} tasks. Classic examples of set-based tasks for a robotic manipulator are the mechanical joint limits, the obstacle avoidance and arm manipulability tasks. In the last years, a great effort has been made in order to extend task-priority frameworks to handle set-based tasks, as for example done in \cite{Simetti_jirs2016}. In particular, the singularity-robust multi-task priority inverse kinematic framework has been extended to handle set-based tasks in \cite{MoeAnt2016frontiers}.

The key idea is that a set-based task can be seen as an equality-based one which gets active or inactive depending on its current value. In particular, it is necessary to set different reference values for each set-based task: physical thresholds $\sigma_{M}$ ($\sigma_{m}$), safety thresholds $\sigma_{s,u}~<~\sigma_M$ ($\sigma_{s,l}~>~\sigma_m$), and activation thresholds $\sigma_{a,u} =  \sigma_{s,u}~-~\epsilon$ ($\sigma_{a,l}~=~\sigma_{s,l}~+~\epsilon$). Figure \ref{fig:thre} shows all the mentioned thresholds. When the task value exceeds an activation threshold, it has to be added to the task hierarchy as a new equality-based task with desired value:
 \begin{equation}
\sigma_d=\left\{\begin{array}{l l}
\sigma_{s,u} &\text{if $\sigma\geq\sigma_{a,u}$}\\
\sigma_{s,l} &\text{if $\sigma\leq\sigma_{a,l}$}
 \end{array} \right.
\end{equation}

\begin{psfrags}
    \psfrag{Active}[1pt][1pt][0.6]{Active}
    \psfrag{Inactive}[1pt][1pt][0.6]{Inactive}
    \psfrag{epsilon}[1pt][1pt][0.8]{$\epsilon$}
    \psfrag{sigmam}[1pt][1pt][0.8]{$\sigma_{m}$}
    \psfrag{sigmaM}[1pt][1pt][0.8]{$\sigma_{M}$}
    \psfrag{sigmaSL}[1pt][1pt][0.8]{$\sigma_{s,l}$}
    \psfrag{sigmaSU}[1pt][1pt][0.8]{$\sigma_{s,u}$}
    \psfrag{sigmaAL}[1pt][1pt][0.8]{$\sigma_{a,l}$}
    \psfrag{sigmaAU}[1pt][1pt][0.8]{$\sigma_{a,u}$}
    \mypsfrag{6}{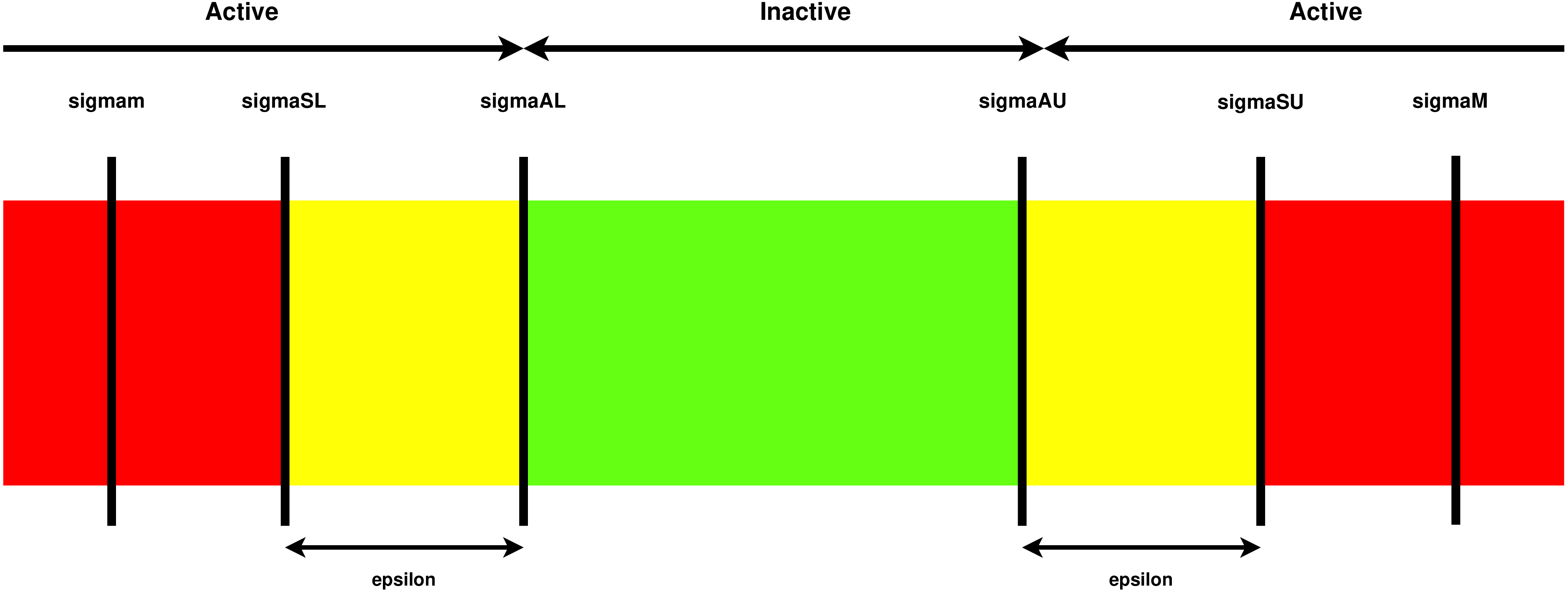}{-12pt}{Activation and physical thresholds of a set-based task}{fig:thre}
\end{psfrags}
Then it can be deactivated when the solution of the hierarchy that contains only the other tasks would push its value toward the valid set. More specifically, it is possible to check whether a generic solution $\dot\bfq$ makes a set-based task $\sigma_A$ go beyond the desired limit or not by evaluating its projection in the task space. Defining $\bfJ_A$ as the Jacobian matrix of $\sigma_A$, if $\bfJ_A \dot\bfq > 0$ the solution would increase the set-based task value, otherwise if $\bfJ_A \dot\bfq < 0$ the solution would decrease it. In this way, $\sigma_A$ can be deactivated if
\begin{equation}\label{eq:cond1}
\sigma_A \geq \sigma_{a,u} \; \land \; \bfJ_A \dot\bfq < 0
\end{equation}
or
\begin{equation}\label{eq:cond2}
\sigma_A \leq \sigma_{a,l} \; \land \; \bfJ_A \dot\bfq > 0
\end{equation}

\subsection{Kinematic singularity handling}
A configuration is defined as kinematically singular when the corresponding jacobian matrix $\bfJ(\bfq)$ loses rank, and it is associated with a loss of mobility of the manipulator's end-effector. The reference velocity in output from the inverse kinematics algorithm diverges, leading the system to instability \cite{chiaverini1994review}. This kind of situations are undesirable and need to be properly handled or avoided in order to guarantee the safety of the operations. The most popular way is to exploit the Damped Least-Square pseudoinverse, defined as:
$$
\bfJ_{\text{DLS}}^\dagger=\bfJ^T(\bfJ \bfJ^T +\lambda^2 \bfI_m)^{-1}
$$
in which a proper choice of the damping factor $\lambda$ can avoid the undesirable effects of the kinematic singularity. In literature there are many different
algorithm for determining $\lambda$ in function of the minimum singular value of the Jacobian matrix and the task error norm. In \cite{DivNatAnt_ifac17} a comparison among different algorithms is carried out. For this work, the dynamic threshold for the damping factor presented in \cite{Baerlocher_phd2001} has been chosen:
$$
\lambda=\begin{cases}
0 & \text{if} \quad \sigma_{\text{min}} \geq \sigma^\star \\
\sqrt{\sigma_{\text{min}} (\sigma^\star - \sigma_{\text{min}})} & \text{if} \quad \sigma^\star/2 \leq \sigma_{\text{min}} < \sigma^\star \\
\sigma^\star/2 & \text{if} \quad \sigma_{\text{min}} < \sigma^\star/2
\end{cases}
$$
with:
$$
\sigma^\star=\frac{||\tilde\bfsigma||}{||\dot\bfq||_{\text{max}}}
$$
where $||\tilde\bfsigma||$ is the task error norm and $||\dot\bfq||_{\text{max}}$ is the maximum joint velocity norm.

\subsection{Implemented tasks}
For the presented system several set-based and equality-based tasks have been implemented.
\begin{itemize}
\item End-effector configuration (equality-based): assign a combination of end-effector position and orientation;
\item Mechanical joints limits (set-based): set thresholds on joints positions;

\item Obstacle avoidance (set-based): make the end-effector of the manipulator keep a minimum distance from a target obstacle;

\item Virtual walls (set-based): keep the end-effector at a minimum distance from a virtual plane;

\end{itemize}
In order to effectively and safely operate the manipulator, it is useful to divide all these tasks in three groups and assign them a priority level \cite{ceriani2015reactive}.
\begin{enumerate}
\item \textbf{Safety related tasks:} this group contains all the safety-related tasks, such as mechanical joint limits, obstacle avoidance and virtual walls.
                                     Since they assure the integrity of the system and of the environment in which it operates, the highest priority level needs to 										 be assigned to them.
\item \textbf{Operational tasks:} this group contains all the tasks aimed at the accomplishment of the mission, such as the end-effector position, orientation or configuration.

\item \textbf{Optimization tasks:} this group contains all those tasks that are not strictly necessary for the effective accomplishment of the operation, but they 										might help in making it in a more efficient way. In this category lie tasks such as the arm manipulability and the field of view. 
\end{enumerate}

\section{Minimum distance evaluation using depth measurements}\label{sec:distances}
We consider a depth sensor used for monitoring the environment in which the robot operates and our goal is to evaluate the distance between a Control Point $\bfP$ and an obstacle point $\bfO$ using their depth space representation. The Depth space is a $2.5$-dimensional space in which the first two coordinates represent the projection of a point in the camera plane and the third coordinate is the distance between the point and the camera. The depth sensor is usually modelled as a pin-hole camera, which is composed by two matrices expressing the intrinsic parameters that model the projection of a point in the image plane, and the extrinsic parameters that represent the transformation between the reference and the sensor:
$$
\boldsymbol{\mathcal{K}} = \begin{pmatrix} f s_x & 0 & c_x \\ 0 & f s_y & c_y \\ 0 & 0 & 1 \end{pmatrix}, \quad \bfepsilon = \begin{pmatrix} \bfR & \bft \\ \bf0 & 1 \end{pmatrix}
$$
where $f$ is the focal length of the camera, $s_x$ and $s_y$ are the pixel dimensions, $c_x$ and $c_y$ are the coordinates of the center of the image plane. Given a control point $\bfP_d = \left[p_x \; p_y \; d_p \right]^T$ in the depth space, its projection in the cartesian sensor space is given by:
$$
P_{s,x} = \frac{(p_x - c_x) d_p}{f s_x}, \quad P_{s,y} = \frac{(p_y - c_y) d_p}{f s_y}, \quad P_{s,z} = d_p 
$$
and its distance vector $\bfV_s = \left[v_{s,x} \; v_{s,y} \; v_{s,z} \right]^T$, expressed in cartesian sensor space,  from an obstacle point $\bfO_d = \left[o_x \; o_y \; d_o \right]^T $ can be computed as:
$$
v_{s,x} = \frac{(o_x - c_x) d_o - (p_x - c_x) d_p}{f s_x}
$$
$$
v_{s,y} = \frac{(o_y - c_y) d_o - (p_y - c_y) d_p}{f s_y}
$$
$$
v_{s,z} = d_o - d_p
$$
For further details about this formulation see \cite{flacco2015depth}.

We are interested in monitoring the environment and in detecting all the obstacle points close to three different control points placed on the manipulator. In particular the control points are placed on the fourth, the sixth and the seventh joint, namely on the elbow, on the wrist and on the hand of the manipulator.
For each one of these control points $\bfP_i$ it is useful to define a {\textit{region of surveillance}} $\bfS_i$ composed by a cube of side $2 \rho$ centered at $\bfP_i$. It is necessary to compute the distances among all the points in the depth image contained in these three regions of surveillance and all the control points and select the closest ones. 

It is important to notice that the manipulator needs to be removed from the depth image, otherwise the obstacle points closest to the chosen control points would always belong to the manipulator itself, and the computed minimum distance would be equal to zero. For this purpose, the Real-Time URDF filter ROS package \cite{urdf_filter} has been used. It receives as input the URDF model of the arm, the joint positions and the transformation between the robot frame and the camera frame and computes the depth image without the manipulator.

Figure \ref{fig:distance2} shows the output of the minimum distance evaluation algorithm. The three control points are placed on the manipulator (light grey circles) and the corresponding minimum distance points (white circles) are computed in real time. Notice that the original meshes of the manipulator have been replaced by larger boxes, in order to avoid irregularities in the manipulator removal procedure.
\begin{psfrags}
    \mypsfrag{6}{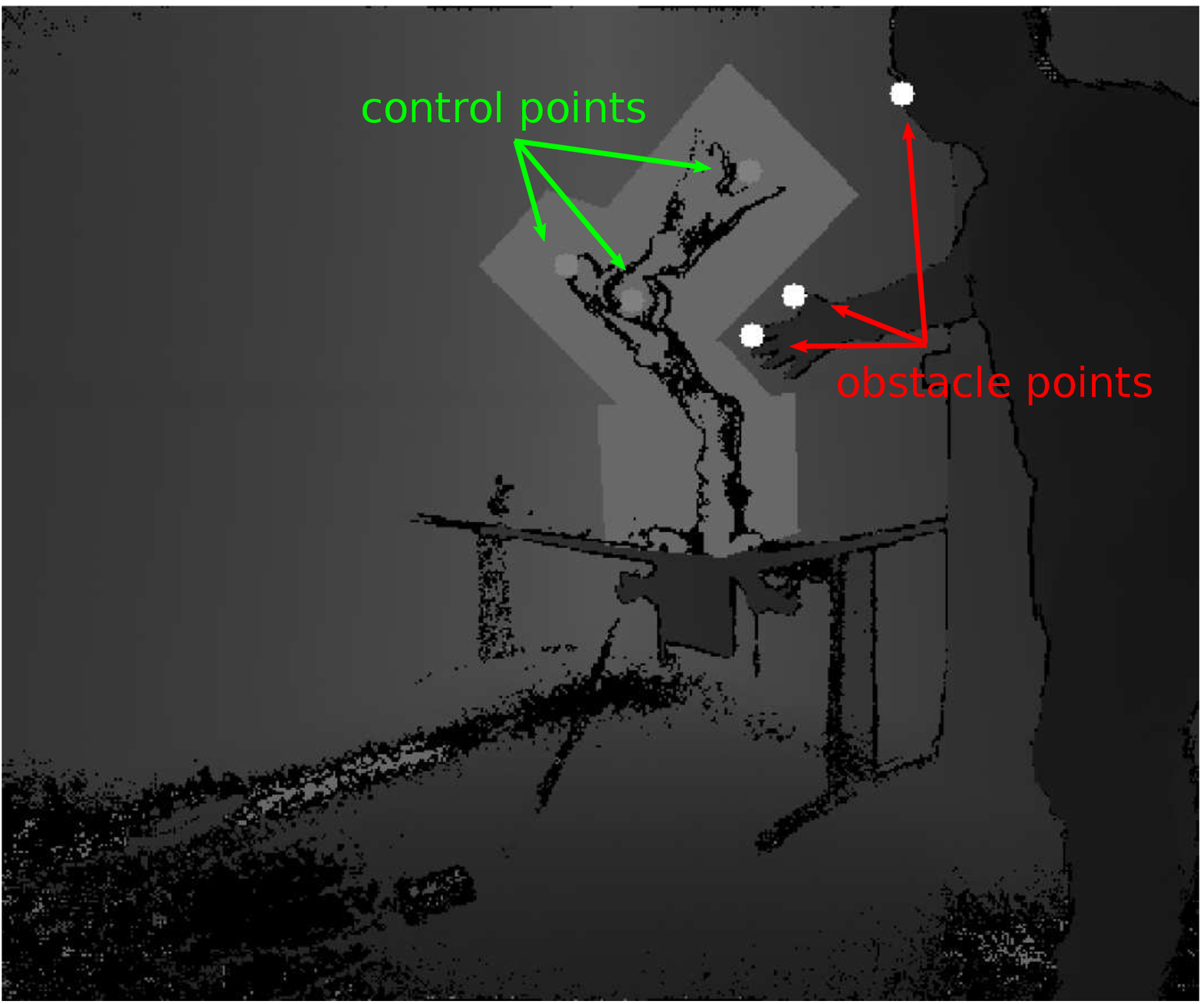}{-12pt}{Minimum distance computation: control points (light grey circles) and corresponding closest points (white circles) )}{fig:distance2}
\end{psfrags}

In the controller three set-based obstacle avoidance tasks are defined, one for each control point. The distance computation algorithm outputs the coordinates of the three points closest to the selected control points, and they are sent to the control algorithm that activates the corresponding task if the computed distance is lower than the chosen activation threshold. The vectors $\bfV_i = \bfO_i - \bfP_i$, being the distance vectors expressed in the arm base frame,  are additionally used for the task jacobians computation. The $i$-th obstacle avoidance task value is:
$$
\sigma_i = \sqrt{(\bfO_{i,min} - \bfP_i)^T \; (\bfO_{i,min} - \bfP_i)} \quad i=1,2,3
$$
where $\bfP_1$, $\bfP_2$ and $\bfP_3$ are the position of the fourth, the sixth and the seventh joint expressed in the arm base frame and $\bfO_{i,min}$ is the corresponding closest point expressed in the arm base frame. The associated Jacobian is computed as:
$$
\bfJ_i = - \frac{(\bfO_{i,min} - \bfP_i)^T}{\sqrt{(\bfO_{i,min} - \bfP_i) (\bfO_{i,min} - \bfP_i)}} \;\bfJ_\text{pos} ^{1..j}
$$
where $\bfJ_\text{pos} ^{1..j} \in \mathbb{R}^{3\times n}$  is the matrix composed by the first $j-1$ columns of the position Jacobian, with $j$ equal to the index of the joint taken as control point, filled with zeros from the column $j$ to $n$.
$$
\bfJ_\text{pos} ^{1..j} = \begin{bmatrix} \bfJ_\text{pos}^{1} & \bfJ_\text{pos}^{2} & \dots & \bfJ_\text{pos}^{j-1} & \bf0 \end{bmatrix}
$$

\section{Experiments}\label{sec:experiments}
\subsection{Experimental setup}

In order to validate the proposed system a number of experiments have been carried out on the Kinova Jaco$^2$ 7 DOF manipulator.
 The arm base frame is labelled with a marker which is detected in real-time using the Aruco library \cite{Aruco2014}. The transformation between the arm base frame and the camera frame is given as input to the real-time URDF filter for removing the arm from the depth image and it is used for the transformation of the closest obstacle points in the arm base frame. The sensor used for the depth image acquisition is a Microsoft Kinect 2 \cite{iai_kinect2}, and the library used for the minimum distance computation is the PCL (Point Cloud Library) \cite{Rusu_ICRA2011_PCL}. The arm is controlled at 100 Hz, while the distance computation algorithm run at 30 Hz, which is the maximum acquisition frequency of the Kinect sensor. 

In the following, the results of two case studies are shown, demonstrating the effectiveness of the developed system.

\subsection{First case study}
In the first case study the following prioritized task hierarchy has been chosen:
\begin{enumerate}
    \item Second, fourth and sixth joints limits
    \item Obstacle avoidance for the three control points
    \item End effector position
\end{enumerate}

The joint limits have been chosen matching their actual mechanical limits, in order to avoid that the manipulator hits its own structure.
The minimum distance from the obstacles for the three control points placed on the manipulator has been set at 35 cm.
The end-effector is asked to sequentially reach two different predefined waypoints $\bfp_d^1 = \left[-0.5 \; 0.4 \; 0.7 \right]^T$ and $\bfp_d^2 = \left[0.5 \; 0.4 \; 0.7 \right]^T$ expressed in the arm base frame. 
Figure \ref{fig:case1-error} shows the position error over time during the experiment, while Fig. \ref{fig:case1-obst} shows the distance between the closest obstacle points and the three control points, together with their minimum thresholds.
\begin{psfrags}
    \mypsfrag{7}{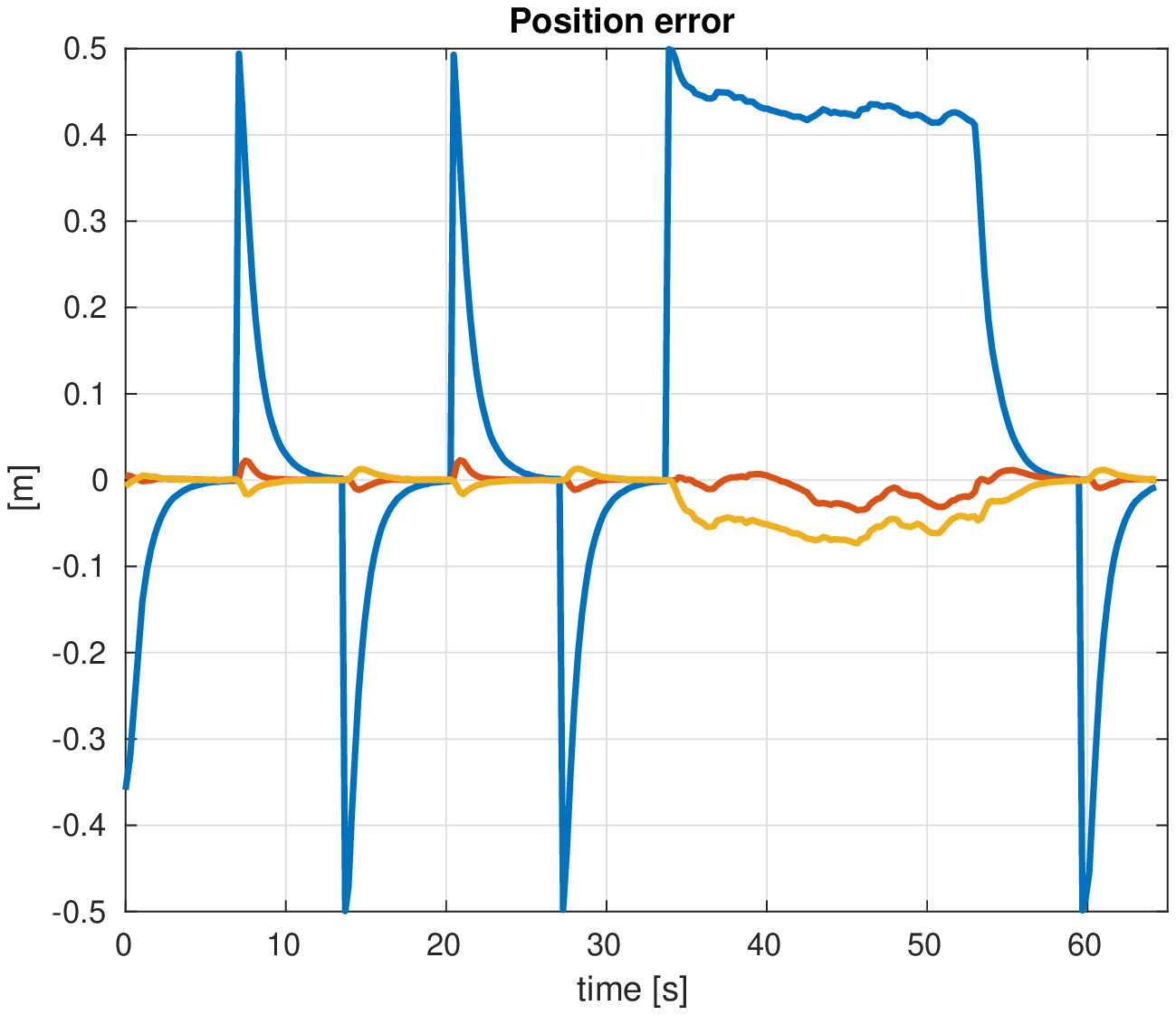}{-12pt}{First case study: position error. From $t=0$s to $t=35$s the arm is free to reach sequentially the two predefined waypoints. From $t=35$s to $t=55$s a person steps into the scene and the obstacle avoidance tasks get activated, stopping the motion of the manipulator. From $t=55$s the person steps away from the manipulator, which is free to continue its movement toward the desired waypoints.}{fig:case1-error}
\end{psfrags}
\begin{psfrags}
    \mypsfrag{7}{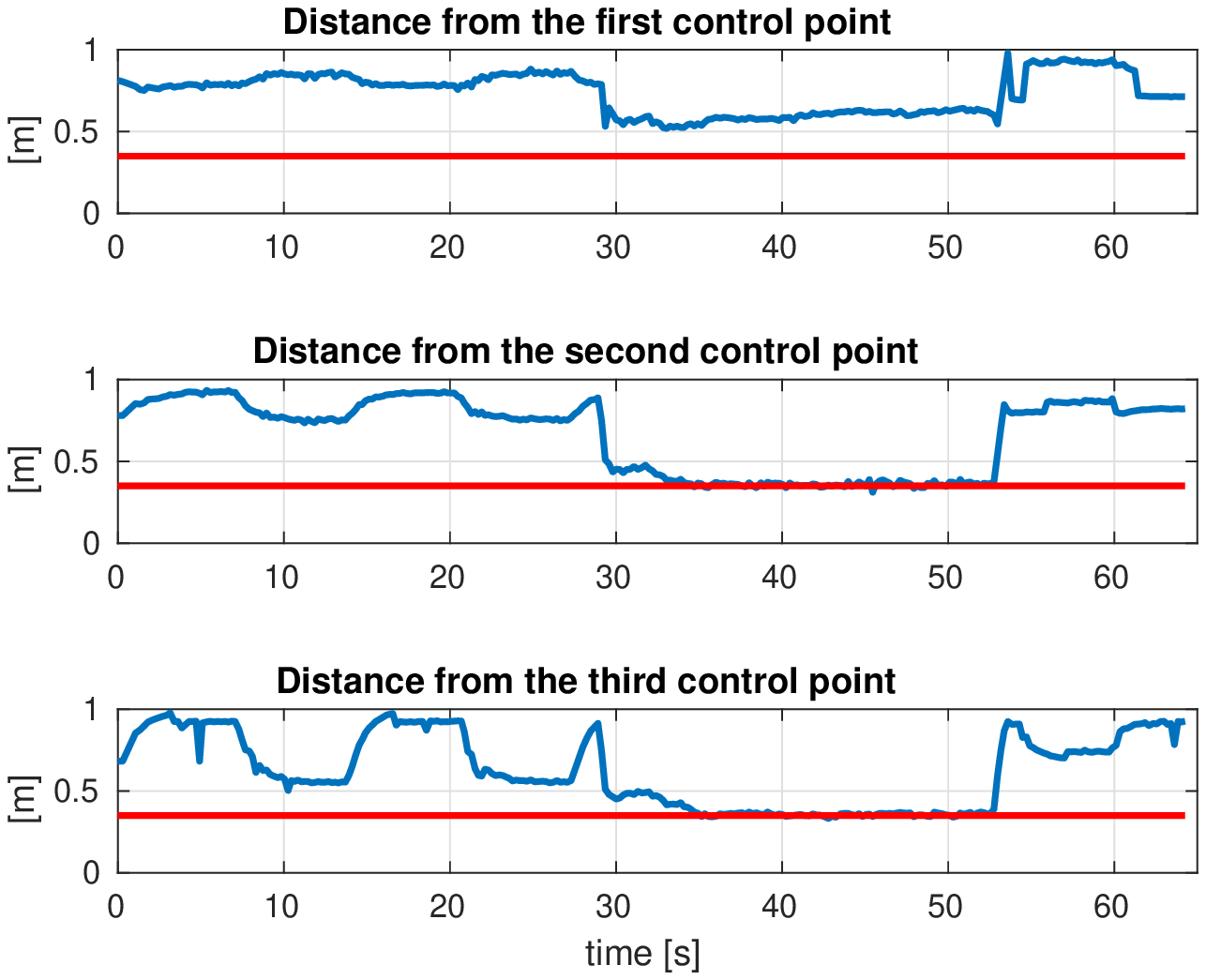}{-22pt}{First case study: distance between the control points and the respective closest obstacle points. The desired minimum thresholds are highlighted (red lines). At $t=35$s a person steps into the scene and the obstacle avoidance tasks for the second and third control points get active, keeping the distance above the chosen thresholds. At $t=55$s the person exits from the scene and the tasks deactivates.}{fig:case1-obst}
\end{psfrags}

At the beginning of the experiment a person stands near the arm, keeping the distance above the activation thresholds of the obstacle avoidance tasks. The arm is free to move reaching sequentially the two predefined waypoints.
At $t=35$s the person steps into the scene, getting closer to the manipulator. Two obstacle avoidance tasks get activated (the ones corresponding to the hand and the wrist), and the control algorithm stops the motion of the manipulator, preventing the collision. Figure \ref{fig:case1-obst} shows that the minimum threshold for the obstacle avoidance tasks is respected, while in Fig. \ref{fig:case1-error} it is clear that the position error is high while the obstacle avoidance tasks are active. 
At $t= 55$s the person steps back, triggering the deactivation of the obstacle avoidance tasks and allowing the manipulator to continue its movement toward the desired waypoints.
Figure \ref{fig:case1-joints} shows the joint values and their upper and lower thresholds during all the experiments. 
\begin{psfrags}
    \mypsfrag{7}{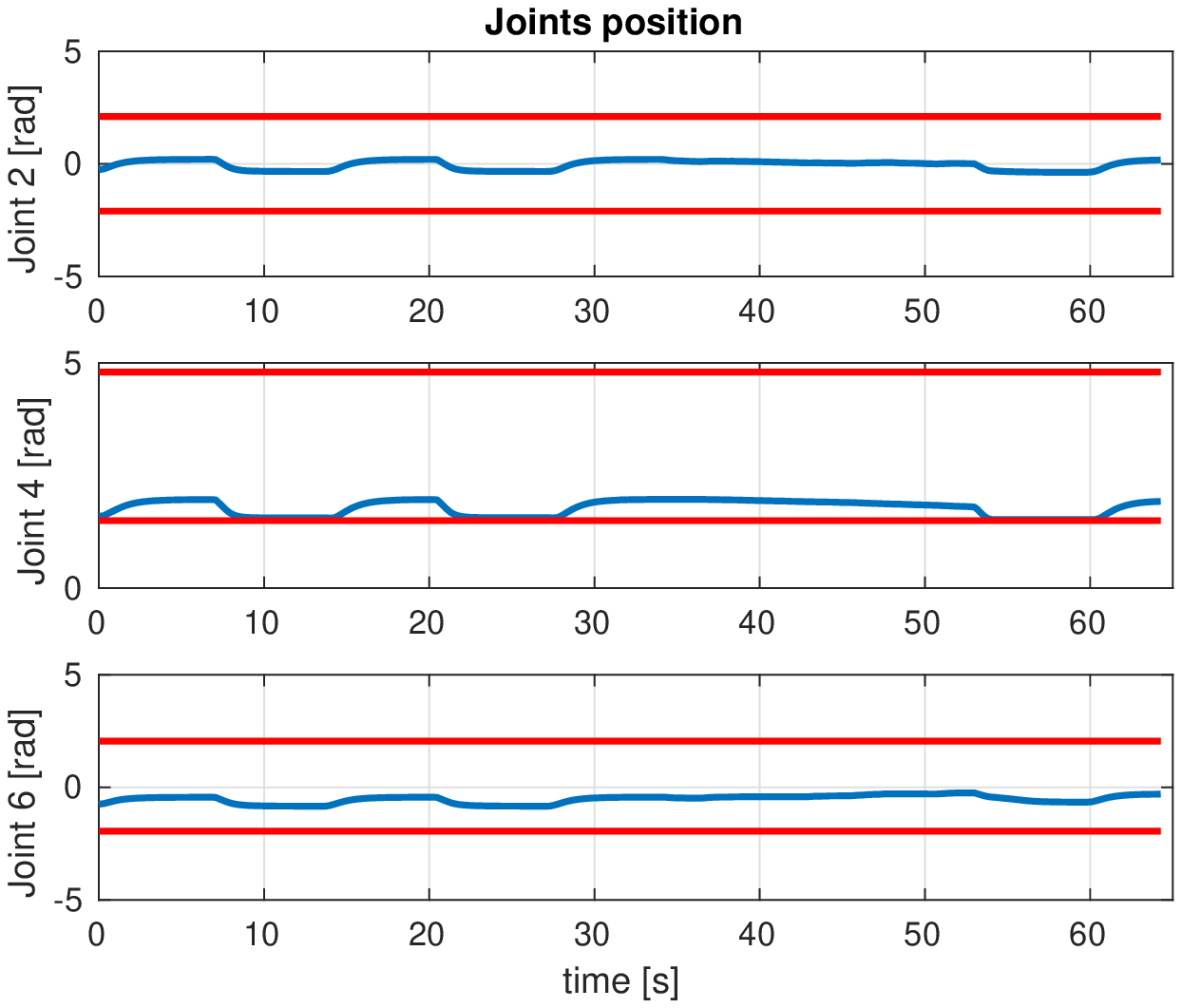}{-18pt}{First case study: second, fourth and sixth joint value (blue line) and minimum/maximum thresholds (red line). The fourth joint limit task gets active during the trajectory and the control algorithm stops its motion respecting the desired threshold. }{fig:case1-joints}
\end{psfrags}
It is worth noticing that the fourth joint task gets active during the trajectory, and the control algorithm stops its motion in order to respect the given threshold. 

\subsection{Second case study}  
For the second case study a more complex task hierarchy has been chosen:
\begin{itemize}

    \item Second, fourth and sixth joints limits
    \item Virtual walls: the end-effector is forced to stay within six virtual walls, creating a virtual box around the manipulator
    \item Obstacle avoidance for the three control points
    \item End effector position
    
\end{itemize}

The end-effector is asked to keep a constant position, while a person tries to touch the control points on the manipulator. When the distances reach the chosen thresholds the obstacle avoidance tasks get active and the arm starts moving in order to avoid the collision with the person. Figure \ref{fig:screen-video} and Fig. \ref{fig:screen-video2} show a sequence of screenshots of the experiment.
\begin{psfrags}
    \mypsfragtwocolumn{14}{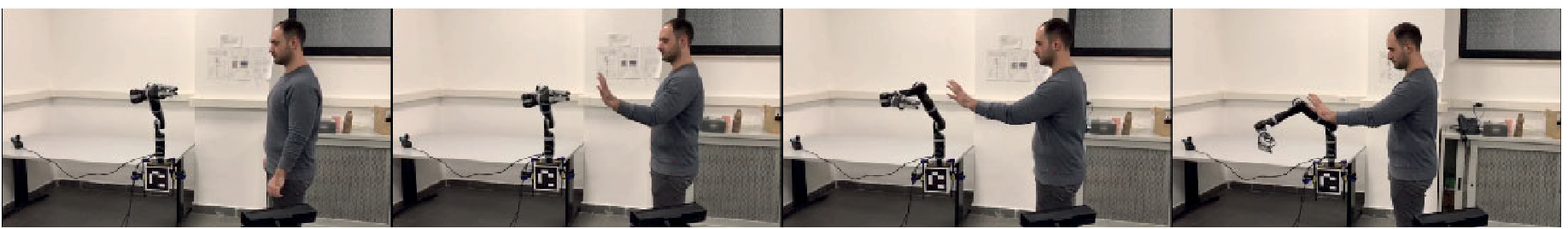}{-12pt}{Screenshots of the experiment. The person tries to touch the end-effector and the arm moves away.}{fig:screen-video}
\end{psfrags}

\begin{psfrags}
    \mypsfragtwocolumn{14}{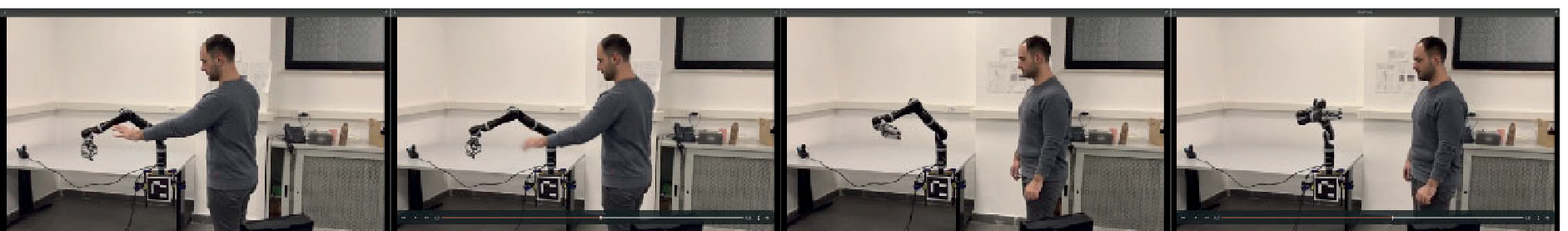}{-12pt}{Screenshots of the experiment. The person steps away from the arm and it returns in the initial position}{fig:screen-video2}
\end{psfrags}
The six virtual walls impose the following limits on the $x$,$y$ and $z$ coordinates in the arm base frame of the end-effector: 
\begin{itemize}
\item $\mathcal{W}_{z,1}$ and $\mathcal{W}_{z,2}$ make the end-effector stay between 0.2 m and 0.9 m on the $z$ axis
\item $\mathcal{W}_{y,1}$ and $\mathcal{W}_{y,2}$ make the end-effector stay between -0.5 m and 0.5 m on the $y$ axis
\item $\mathcal{W}_{x,1}$ and $\mathcal{W}_{x,2}$ make the end-effector stay between -0.5 m and 0.5 m on the $x$ axis 
\end{itemize}
These thresholds have been chosen in order to avoid that the arm reaches the boundary of its workspace, thus the corresponding singular configuration. Additionally the limit on the $z$ coordinate prevents the end-effector to hit the table it is attached on.
Figure \ref{fig:case2-walls} shows the  end-effector position over time, together with the limits imposed by the virtual walls. The person tries to {\textit{push}} the end-effector beyond the walls but the associated tasks get active, stopping the motion in that direction at the chosen thresholds.  
\begin{psfrags}
\mypsfrag{7}{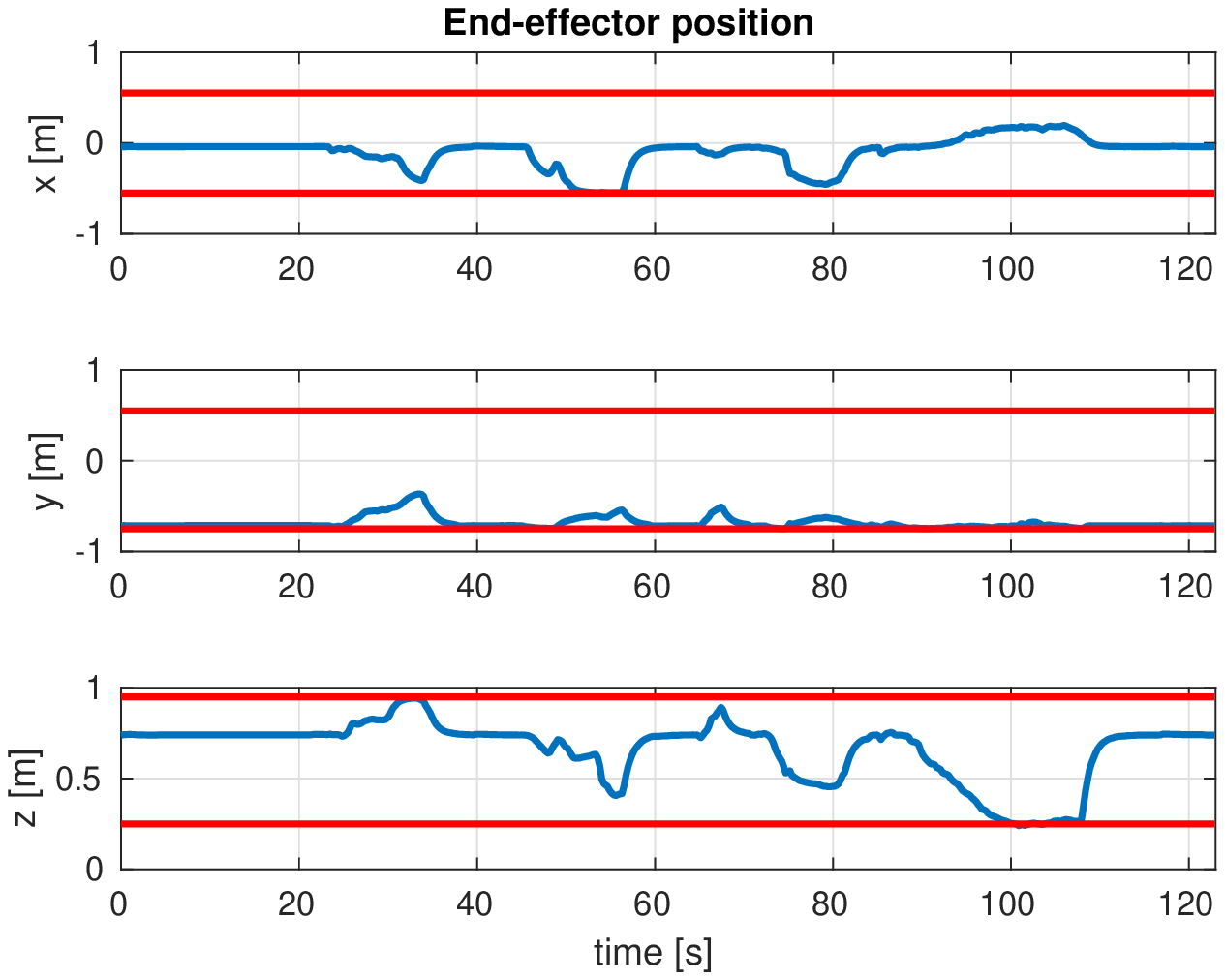}{-22pt}{Second case study: end-effector $x$ (top), $y$ (middle) and $z$ (bottom) coordinates (blue) and thresholds imposed by the virtual walls (red). The end-effector stays within the chosen box even if the person tries to push it beyond the virtual walls.}{fig:case2-walls}
\end{psfrags}

During all the experiment the minimum distance between the person and the three control points on the manipulator is kept above the chosen thresholds, as shown in Fig. \ref{fig:case2-obst}. Notice that the chattering phenomenon is due to the fact that the person is moving and the points at minimum distance change over time.
\begin{psfrags}
    \mypsfrag{7}{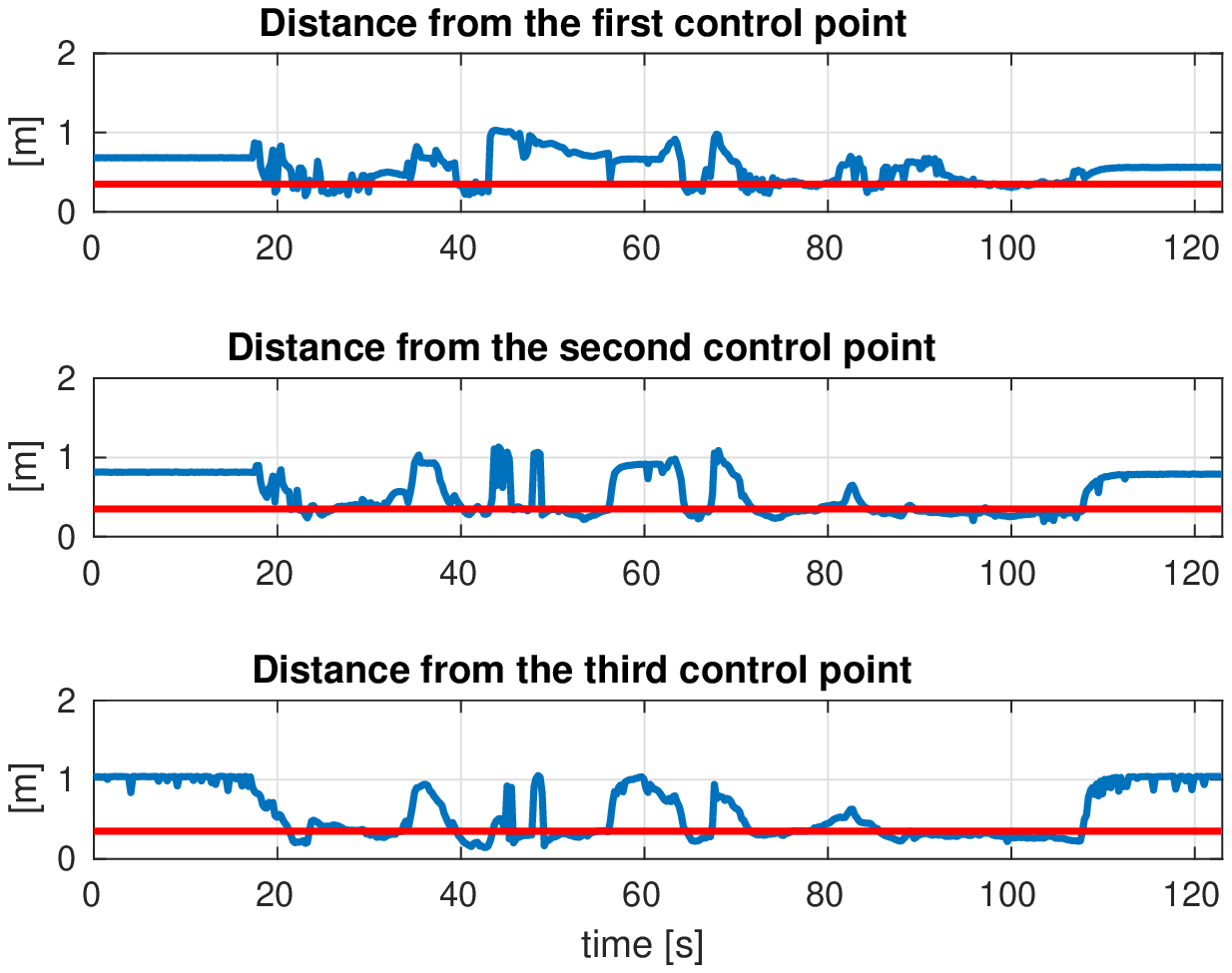}{-22pt}{Second case study: distance from the three control points (blue) and corresponding minimum thresholds (red). The arm tries to keep the distance beyond the chosen limits. }{fig:case2-obst}
\end{psfrags}
Finally, Fig. \ref{fig:case2-joints} shows the joint values with their corresponding limits. It is clear the their positions never exceeds the imposed limits during all the experiment.
\begin{psfrags}
    \mypsfrag{7}{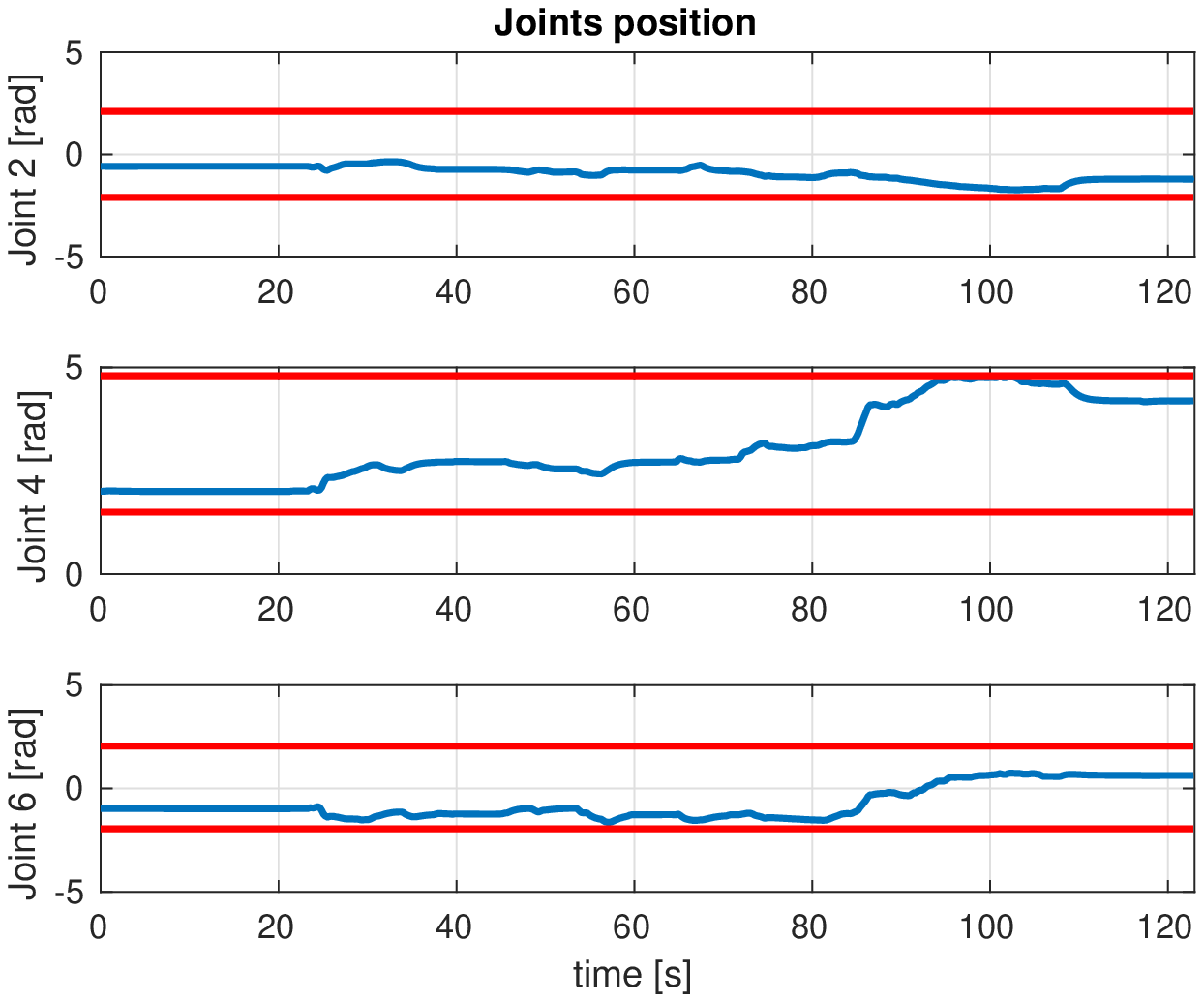}{-16pt}{Second case study: joint positions (blue) and corresponding threshodls (red)}{fig:case2-joints}
\end{psfrags}

\section{Conclusions and future work}\label{sec:conc}
In this paper we presented a system that allows safety operations with a robotic manipulator. The control algorithm that handles a task hierarchy has been explained, focusing the attention on the obstacle avoidance, the joint limits and the virtual walls tasks. The algorithm for the real-time evaluation of the closest obstacles to the manipulator from depth measurements has been described, together with its integration in the motion controller. Finally experimental results on a 7 DOF Kinova Jaco$^2$ using a Kinect sensor for the obstacles detection have been shown, proving the effectiveness of the developed system.

Further efforts will be used in two main directions. First of all we want to improve the obstacles detection phase by using multiple Kinect sensors, increasing the field of view of the overall system and minimizing occlusions issues. The second direction will be making the system robust with respect to the occurring in local minima problems, that are very likely in gradient-based methods. The idea would be to integrate in the framework a motion planner, capable of detecting such situations and of replanning the motion of the system.
\section*{Acknowledgments}
This work was supported by the European Community through the projects EUMR (H2020-731103-2), ROBUST (H2020-690416), DexROV (H2020-635491) and AEROARMS(H2020-635491).
\bibliography{biblio}

\end{document}

%% file: bmit2e.tex
\font\bfmath=cmmib10
\mathchardef\Gamma="7100
\mathchardef\Delta="7101
\mathchardef\Theta="7102
\mathchardef\Lambda="7103
\mathchardef\Xi="7104
\mathchardef\Pi="7105
\mathchardef\Sigma="7106
\mathchardef\Upsilon="7107
\mathchardef\Phi="7108
\mathchardef\Psi="7109
\mathchardef\Omega="710A

\mathchardef\alpha="710B
\mathchardef\beta="710C
\mathchardef\gamma="710D
\mathchardef\delta="710E
\mathchardef\epsilon="710F
\mathchardef\zeta="7110
\mathchardef\eta="7111
\mathchardef\theta="7112
\mathchardef\iota="7113
\mathchardef\kappa="7114
\mathchardef\lambda="7115
\mathchardef\mu="7116
\mathchardef\nu="7117
\mathchardef\xi="7118
\mathchardef\pi="7119
\mathchardef\rho="711A
\mathchardef\sigma="711B
\mathchardef\tau="711C
\mathchardef\upsilon="711D
\mathchardef\phi="711E
\mathchardef\chi="711F
\mathchardef\psi="7120
\mathchardef\omega="7121
\mathchardef\epsilon="7122

\mathchardef\varepsilon="7122
\mathchardef\vartheta="7123
\mathchardef\varpi="7124
\mathchardef\varrho="7125
\mathchardef\varsigma="7126
\mathchardef\varphi="7127
\mathchardef\imath="717B
\mathchardef\jmath="717C

\def\bfp{{\mbox{\boldmath $p$}}}
\def\bfq{{\mbox{\boldmath $q$}}}

\def\bft{{\mbox{\boldmath $t$}}}

\def\bfI{{\mbox{\boldmath $I$}}}
\def\bfJ{{\mbox{\boldmath $J$}}}
\def\bfK{{\mbox{\boldmath $K$}}}

\def\bfN{{\mbox{\boldmath $N$}}}
\def\bfO{{\mbox{\boldmath $O$}}}
\def\bfP{{\mbox{\boldmath $P$}}}

\def\bfR{{\mbox{\boldmath $R$}}}
\def\bfS{{\mbox{\boldmath $S$}}}

\def\bfV{{\mbox{\boldmath $V$}}}

\def\bfepsilon{{\mbox{\boldmath $\epsilon$}}}

\def\bfsigma{{\mbox{\boldmath $\sigma$}}}

%% file: mymacro.tex
\def\smallbfW{{\raise1.5pt\hbox{\mbox{\boldmath $_W$}}}}

\def\mypsfrag#1#2#3#4#5{
        \begin{figure}[htp]
           \begin{center}
              {\leavevmode
                 {\includegraphics[width=#1truecm]{#2.eps}}
              }
           \end{center}
           \vspace{#3}
           \caption{#4}
           \vspace{-10pt}
           \label{#5}
        \end{figure}
}

\def\mypsfragtwocolumn#1#2#3#4#5{
        \begin{figure*}[htp]
           \begin{center}
              {\leavevmode
                 {\includegraphics[width=#1truecm]{#2.eps}}
              }
           \end{center}
           \vspace{#3}
           \caption{#4}
           \label{#5}
        \end{figure*}
}

\def\my4psfrag#1#2#3#4#5#6#7#8{
        \begin{figure}[htp]
        \begin{center}
	        \begin{tabular}[h]{c c}
              {\leavevmode{\includegraphics[width=#1truecm]{#2.eps}}}
              &
              {\leavevmode{\includegraphics[width=#1truecm]{#3.eps}}} \\
              {\leavevmode{\includegraphics[width=#1truecm]{#4.eps}}}
              &
              {\leavevmode{\includegraphics[width=#1truecm]{#5.eps}}}
   	     \end{tabular}
           \vspace{#6}
           \caption{#7}
           \label{#8}
        \end{center}
        \end{figure}
}

\def\mydouble4psfrag#1#2#3#4#5#6#7#8{
        \begin{figure*}[htp]
        \begin{center}
            \begin{tabular}[h]{c c}
              {\leavevmode{\includegraphics[width=#1truecm]{#2.eps}}}
              &
              {\leavevmode{\includegraphics[width=#1truecm]{#3.eps}}} \\
              {\leavevmode{\includegraphics[width=#1truecm]{#4.eps}}}
              &
              {\leavevmode{\includegraphics[width=#1truecm]{#5.eps}}}
         \end{tabular}
           \vspace{#6}
           \caption{#7}
           \label{#8}
        \end{center}
        \end{figure*}
}